# A real-time vehicles detection algorithm for vision-based sensors

Bartłomiej Płaczek

Faculty of Transport, Silesian University of Technology,
ul. Krasinskiego 8, 40-019 Katowice, Poland
bartlomiej.placzek@polsl.pl

**Abstract.** A vehicle detection plays an important role in the traffic control at signalised intersections. This paper introduces a vision-based algorithm for vehicles presence recognition in detection zones. The algorithm uses linguistic variables to evaluate local attributes of an input image. The image attributes are categorised as vehicle, background or unknown features. Experimental results on complex traffic scenes show that the proposed algorithm is effective for a real-time vehicles detection.

## 1   Introduction

The information concerning the presence of vehicles in the predefined detection zones on traffic lanes is essential for the adaptive traffic control at signalised intersections. This information can be effectively acquired using vision-based vehicle detectors. The idea is not new. However, fast and robust image processing algorithms are still being sought to provide low cost vehicle video-detectors and to improve their performance.

In recent years many vision-based algorithms have been developed for the vehicle detection. Most of them have been designed to perform image segmentation and categorise all pixels of the input image as pixels of vehicles, pixels of background, shadows, etc. [1, 5, 6]. The background subtraction is one of the most popular methods. Several adaptive background models have been proposed for the traffic images segmentation [9, 10]. Different methods have been intended for the counting of vehicles when passing through so-called virtual loops. In [7], the statistical models have been applied to extract image features that allow us to categorise each state of the detector into three categories (road, vehicle head and body) and to recognise passing vehicles. The methods proposed in [2, 8] use the time-spatial images analysis for the task of road traffic flow count. Also 3-D shape models have been dedicated for the vehicles detection [3, 4].

The paper introduces a fast algorithm for the vehicles presence recognition in detection zones. It operates on local image attributes instead of the particular pixels. The image attributes are evaluated for small image regions by using linguistic variables and the fuzzy sets theory. The recognition of vehicles is performed by using simple statistics on the values of the image attributes registered in a video sequence. The occupancy of detection zones is then determined by counting the recognised vehicles features in appropriate image regions. The algorithm was applied for gray-scale video sequences. However, it can be easily extended to colour images. The robustness of the proposed algorithm was verified through extensive testing in various situations. The experimental results reported in this paper proves the real-time performance of the introduced algorithm.

## 2 Image attributes

The pixels intensities and their spatial differences are the low-level image characteristics that can be effectively utilised for the vehicles detection. As the grey-scale images are taken into consideration, the colour attribute was introduced to categorise pixels into three classes: black (with low intensities), grey (medium intensities) and white (high intensities). This colour attribute allows to recognise some features of objects that are visible in traffic scenes, e. g.: black regions of image that usually correspond with bottoms of vehicles in daylight conditions, vehicles headlights having white colour, especially in night images and regions of grey colour which is typical of the road pavement.

The spatial differences of intensities allow us to evaluate contrast between the adjacent image regions. The fact that a high contrast correspond with sharp edges of a vehicle body, is particularly important for the task of vehicles detection. Therefore, the proposed method uses the contrast measure as an relevant image attribute. The applied contrast description of two neighbouring image regions $X$ and $Y$ is based on the distinction of three cases: (1) $X$ is darker than $Y$, (2) $X$ and $Y$ have similar intensities, (3) $X$ is brighter than $Y$.

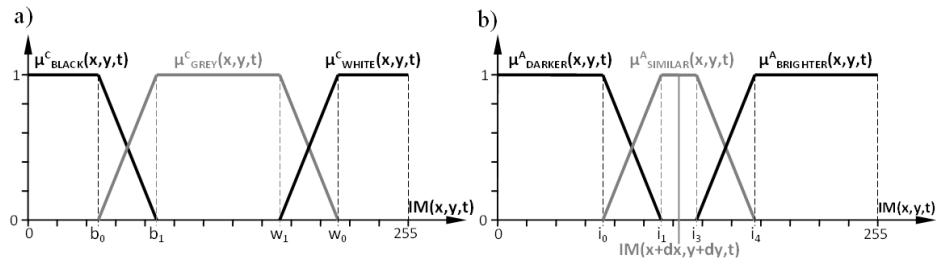

**Fig. 1.** Fuzzy sets definitions for linguistic variables: a) $C$, b) $UR$, $UL$, $LL$ and $LR$.

The local attributes of $t$-th image for coordinates $(x, y)$ are determined taking into account 7x7 square neighbourhood centred on a pixel $(x, y)$. The set of attributes for the pixel $(x, y)$ consists of one colour attribute $C$ and four contrast attributes: $UR$, $UL$,

*LL* and *LR*. The contrast attributes describe the intensity differences between the centre of the analysed image region and its corners: upper right, upper left, lower left and lower right corner respectively. The values of the image attributes are determined by using linguistic variables. The sets of admissible values (linguistic terms) for the introduced attributes are defined as follows:

$$V_C = \{black, grey, white\}, V_{UR} = V_{UL} = V_{LL} = V_{LR} = \{darker, similar, brighter\}. \quad (1)$$

Figure 1 presents the membership functions of fuzzy sets that are attached to the linguistic terms. These functions map values of pixels intensity onto membership degrees of linguistic terms. Membership degrees are computed taking into account the mean values of pixels intensities *IM*(*x*, *y*, *t*) determined for the image regions of size 3x3.

The membership functions of linguistic terms describing the colour of a pixel are depicted in figure 1 a). Parameters $b_0$, $b_1$ and $w_1$, $w_2$ of these functions are calculated by means of histogram analysis performed on the input image. Figure 1 b) shows membership functions defined for terms that are used to describe contrast between two adjacent 3x3 image regions. The first of them is centred on the pixel (*x*, *y*) and the second one is centred on the pixel (*x* + *dx*, *y* + *dy*). The selection of *dx* and *dy* values depends on the type of attribute: *dx* = 2, *dy* = -2 for attribute *UR*, *dx* = -2, *dy* = -2 for *UL*, *dx* = -2, *dy* = 2 for *LL* and finally *dx* = 2, *dy* = 2 for *LR*. Parameters $i_0 \ldots i_4$ of the membership functions are computed by a simple addition of predetermined offsets to the intensity value *IM*(*x* + *dx*, *y* + *dy*, *t*).

As a result of a particular attribute *A* evaluation for the pixel (*x*, *y*) in *t*-th image a triple of membership degrees is computed:

$$A(x, y, t) = [\mu_{a_1}^A(x, y, t), \mu_{a_2}^A(x, y, t), \mu_{a_3}^A(x, y, t)], \quad (2)$$

where $\mu_a^A$ is the membership function of term *a* defined for attribute *A*, $a \in V_A$, $V_A = \{a_1, a_2, a_3\}$ is the set of values (terms) for attribute *A* and $A \in \{C, UR, UL, LL, LR\}$.

## 3  Vehicles features

As it was discussed in the previous section, the values of image attributes are represented by the triples of membership degrees. The proposed algorithm classifies each value of image attribute as a feature of vehicles, a feature of background or an unknown feature. The classification procedure is based on the information regarding the number of occurrences of particular attribute values in images that were previously analysed. It was assumed that the background features appear more frequently in the sequence of images than vehicles features do. Thus, a given attribute value can be recognised as a feature of vehicles if it has the low number of occurrences. When the occurrences number for a given value is high, it can be considered as a background feature. The attribute value will be categorised as an unknown feature if its number of occurrences is medium.

The accumulators arrays were applied in the introduced method to collect the occurrence information for respective values of image attributes. This information is necessary for the recognition of vehicles features in a video sequence. The accumulators arrays are updated after attributes evaluation for each image in the video sequence. Let $AC_a^A(x, y, t)$ denotes an accumulator that corresponds with value (term) $a$ of attribute $A$ evaluated for pixel $(x, y)$ in $t$-th image. The count in the accumulator is incremented, if the value of attribute $A$ for the analysed pixel conforms to the linguistic term $a$. This rule can be expressed by using the following equation:

$$AC_a^A(x,y,t) = AC_a^A(x,y,t-1) + \mu_a^A(x,y,t). \tag{3}$$

The accumulator count is decremented in the opposite situation, i. e. when the current value of attribute $A$ is not $a$:

$$AC_a^A(x,y,t) = AC_a^A(x,y,t-1) - \mu_{\bar{a}}^A(x,y,t). \tag{4}$$

When the standard definition of fuzzy set complement is taken into consideration, the membership degree of $\bar{a}$ is calculated as follows:

$$\mu_{\bar{a}}^A(x,y,t) = 1 - \mu_a^A(x,y,t). \tag{5}$$

Therefore, equations (3) and (4) can be merged to give one formula for the accumulator update operation:

$$AC_a^A(x,y,t) = AC_a^A(x,y,t-1) + 2\mu_a^A(x,y,t) - 1. \tag{6}$$

At the beginning, when the video sequence processing starts, there is no information available on the occurrences of particular attributes values. The accumulators are initialised at zero: $AC_a^A(x, y, 0) = 0$. A low absolute value of the count in an accumulator correspond with insufficient information, which does not allow to categorise the image features unambiguously. The features classification is possible only when the absolute value of the counts in accumulators is appropriately high. Moreover, the vehicle features are recognised if relevant accumulators counts are negative and background features are recognised in case of positive counts.

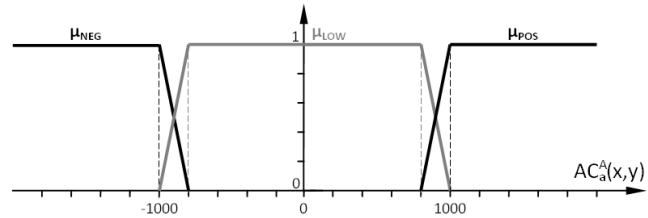

**Fig. 2.** Fuzzy sets definitions for linguistic terms: negative, low and positive.

The terms used above to describe accumulators counts (low, negative, positive) have their own membership functions (fig. 2). The introduced membership functions are used to categorise image features. Fuzzy reasoning method with product t-norm and maximum t-conorm was applied to perform the features classification task. For an recorded image attribute $A(x, y, t)$ the degree of its membership in the vehicle features class ($VF$) is calculated as follows:

$$\mu_{VF}^{A}(x,y,t) = \max_{a}\{\mu_{neg}[AC_{a}^{A}(x,y,t)] \cdot \mu_{a}^{A}(x,y,t)\} \,. \tag{7}$$

When the classes of background features (*BF*) or unknown features (*UF*) are taken into consideration, the membership degrees can be similarly computed. The only difference is in use of $\mu_{pos}$ or $\mu_{low}$ functions instead of $\mu_{neg}$ for *BF* and *UF* classes respectively.

## 4 Occupancy of a detection zone

The task of the proposed algorithm is to provide information on an occupancy of each detection zone. The occupancy $O_D(t)$ is a binary property of the detection zone *D*, determined for *t*-th image in a video sequence. The values of occupancy have the following interpretation: $O_D(t) = 1$ indicates that there is a vehicle occupying zone *D* in *t*-th image and $O_D(t) = 0$ denotes that the detection zone is empty. The procedure of the occupancy determination is based on vehicle features counting in detection zones. The underlying concept is that the number of vehicle features is high when the detection zone is occupied. The number of vehicle features recognised in a detection zone *D* for *t*-th image is calculated according to the following formula:

$$S_D(t) = \sum_{(x,y) \in D} \sum_{A} \mu_{VF}^{A}(x,y,t) \,. \tag{8}$$

In the next step the resulting quantity is thresholded to determine a binary value of occupancy. The hysteresis thresholding method is applied here with two threshold values. The high threshold is used to formulate a condition of detection zone activation (switching occupancy to 1). The low threshold is taken into account when the detection zone is deactivated (occupancy changes from 1 to 0). Let $T_D^{H}(t)$ and $T_D^{L}(t)$ denote high and low threshold for detection zone *D* in the *t*-th image respectively. The occupancy is determined as follows:

$$O_D(t) = \begin{cases} 1, & S_D(t) \geq T_D^{H}(t) \\ 0, & S_D(t) \leq T_D^{L}(t) \\ O_D(t-1), & else \end{cases} \tag{9}$$

The appropriate determination of the thresholds values is crucial for the effective vehicles detection. Constant thresholds cannot ensure a good detection performance because significant variations of $S_D(t)$ range are often encountered for traffic video sequences. The variations are caused by the changes of ambient lighting conditions, weather phenomena, camera vibrations, accuracy limitations of the vehicles features recognition, etc. For the discussed algorithm an adaptive method of threshold determination was introduced to deal with the aforementioned problems. The thresholds values are determined for each image in the analysed video sequence. The following computations are performed to calculate them:

$$T_D^H(t) = \max\{p_D S_D^{MAX}(t) + (1-p_D)S_D^{MIN}(t), 100 p_D\}, \quad T_D^L(t) = \alpha T_D^H(t), \quad (10)$$

where $p_D \in [0; 1]$ is the configuration parameter, which allows to adjust "sensibility" of the detection zone $D$. The current range of $S_D(t)$ values is determined by the interval $[S_D^{MIN}(t); S_D^{MAX}(t)]$. The endpoints of this interval are calculated accordingly:

$$S_D^{MIN}(t) = \begin{cases} S_D^{MIN}(t-1) + \beta^{MIN}, & S_D(t) > S_D^{MIN}(t-1) \\ S_D(t), & S_D(t) \leq S_D^{MIN}(t-1) \end{cases}$$

$$S_D^{MAX}(t) = \begin{cases} S_D^{MAX}(t-1) - \beta^{MAX}, & S_D(t) < S_D^{MAX}(t-1) \\ S_D(t), & S_D(t) \geq S_D^{MAX}(t-1) \end{cases} \quad (11)$$

It was experimentally verified that for $\alpha = 0.8$, $\beta^{MIN} = 0.1$ and $\beta^{MAX} = 0.01$ the algorithm gives the correct results. In most cases the sensibility parameter $p_D$ was set to 0.2. Higher values need to be used when visible edges exist in the background of a detection zone.

Fig. 3 illustrates the occupancy determination procedure for a single detection zone. A sequence of 1150 images was analysed in this example (46 seconds) and eight vehicles were detected. The registered numbers of vehicle features along with thresholds values are presented in the upper chart. The lower plot represents corresponding binary values of the detection zone occupancy.

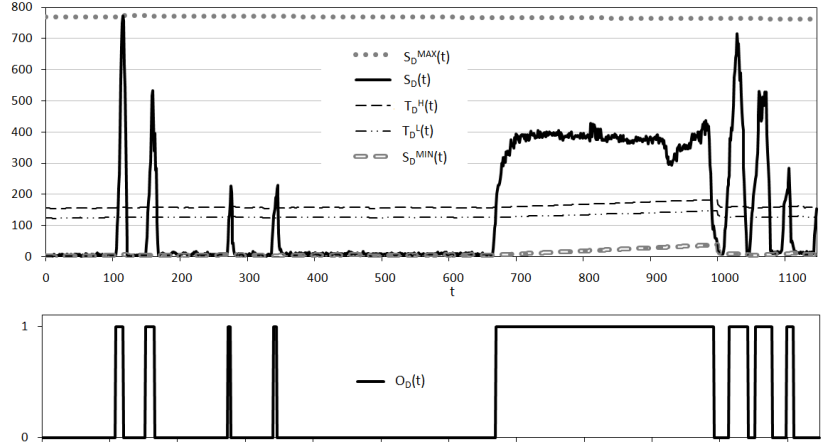

**Fig. 3.** An example of occupancy determination for a single detection zone.

The basic method of occupancy determination, using thresholds defined above, was experimentally tested and some modifications were formulated to improve the algorithm performance. The first modification was motivated by a simple observation that the occupancy value for a detection zone may be changed only when a vehicle is entering into the zone or it is leaving the zone. In both cases the movement of the vehicle can be recognised by video sequence analysis. Thus, an additional condition was introduced for the occupancy determination, which takes into account the results of the movement detection. This modification was found to allow for errors reduction of the vehicles detection, especially when vehicles stops in detection areas for long time-periods. The movement detection was implemented using well-known differential method, based on subtraction of successive images in the video sequence.

The further improvement of the vehicles detection was achieved by occupancy-dependent actualisation of the accumulators. The aim of this modification is to prevent vehicles features registering as the features of image background within occupied detection zones. According to this modification, when the detection zone is occupied, the accumulators actualisation is executed in a different way than that described in section 3. For each pixel ($x$, $y$) in the occupied detection zone $D$ the accumulator $AC_a^A(x, y, t)$ is updated as defined by (6) only if the current count of this accumulator is not low. The update is skipped in an opposite situation. The term "low" in the above condition is interpreted by using a membership function of the fuzzy set presented in fig. 2.

## 5   Experimental results

The proposed algorithm was tested on video sequences captured at several crossroads in various lighting and weather conditions. The summarised length of all tested sequences was about 20 hours. They include day and night time scenes of the congested as well as free traffic flow. The test conditions were complicated with rain, snow, slight camera vibrations and significant light reflections. Fig. 4 shows some examples of the analysed video with the displayed boundaries of detection zones. The size of experimental images was 768 x 512 pixels with 8-bit greyscale.

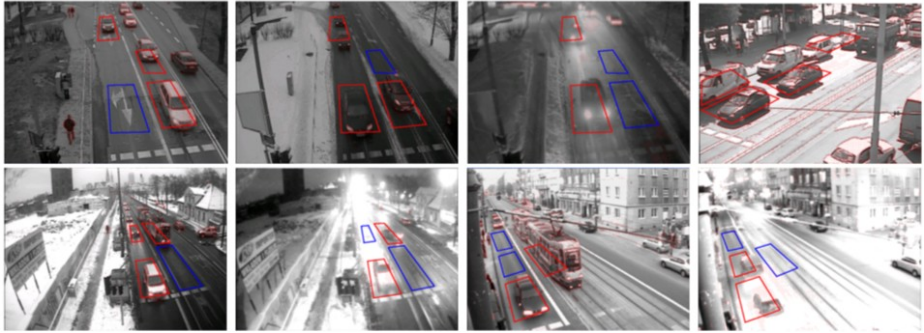

**Fig. 4.** Sample images of video sequences used for the experiments.

A non-optimized implementation of the algorithm achieved processing speed of 30 fps on a Windows PC with Intel Core 2 2,5 GHz. Thus, it fulfils the real-time performance requirement for the video streams operating at 25 fps frame rate.

The accuracy of the vehicles detection was evaluated by the results verification for each time interval of 2 seconds. Two categories of errors were taken into account: a false negative detection was registered if a vehicle was present in a detection zone and the occupancy remained zero during all the time interval. When the detection zone was empty and the occupancy value was one for the time interval then the false positive detection was recognised. The error rate of vehicles detection was lower than 1% for sequences recorded by the cameras installed directly above traffic lanes, in good weather during the day time, when the lighting conditions were stable. This result was obtained also for congested traffic, when vehicles queues were observed. In the worst case, for a night time video sequence with very poor lighting and reflections from the pavement, the error rate rose to 11%.

# 6  Conclusions

The experimental results demonstrate that the proposed algorithm is feasible and promising for the applications in the vision-based vehicle detection. It fulfils the real-time requirement and provides the robust vehicle detection under the complex conditions like lighting transitions, traffic congestion, vehicles headlight impact, etc. The accuracy of detection can be further enhanced by extending the algorithm to colour images. A hardware implementation of this algorithm is possible due to its low computational complexity. It will enable the vision sensors to be more cost-effective in the construction of the vehicles detection systems for road traffic control purposes.

# References

1. Kim, Z., Malik, J.: Fast vehicle detection with probabilistic feature grouping and its application to vehicle tracking. Proc. Int. Conf. Comp. Vision, pp. 524--531. IEEE (2003)
2. Liu, A., Yang, Z.: Video Vehicle Detection Algorithm through Spatio-Temporal Slices Processing. Proc. Int. Conf. Mechatronic Embedded Systems, pp. 1--5. IEEE (2006)
3. Płaczek, B., Staniek, M.: Model Based Vehicle Extraction and Tracking for Road Traffic Control. In: Kurzyński M. et al. (eds.) Advances in Soft Computing. Computer Recognition Systems 2, pp. 844--851. Springer-Verlag, Berlin Heidelberg (2007)
4. Płaczek, B.: Vehicles Recognition Using Fuzzy Descriptors of Image Segments. In: Kurzyński M. et al. (eds.) Advances in Soft Computing. Computer Recognition Systems 3, pp. 79-86. Springer-Verlag, Berlin Heidelberg (2009)
5. Wang, Y., Ye, G.: Joint random fields for moving vehicle detection. Proc. British Machine Vision Conf., vol. 1, pp. 13--22. Leeds, UK (2008)
6. Xu, S., Zhao, Y., Yu, C., Shen, L.: Vehicle Detection Algorithm Based on Shadow Feature. 2008 ISECS Int. Colloquium on Computing, Communication, Control, and Management, Vol. 1, pp. 105--109. IEEE (2008)
7. Yin, M., Zhang, H., Meng, H., Wang, X.: An HMM-Based Algorithm for Vehicle Detection in Congested Traffic Situations. Proc. ITS Conf., pp. 736--741. IEEE (2007)
8. Yue, Y.: A Traffic-Flow Parameters Evaluation Approach Based on Urban Road Video. Int. Journal of Intelligent Engineering and Systems, Vol. 2, No. 1, pp. 33--39, INASS (2009)

9. Zhang, G., Avery, R.P., Wang, Y.: Video-based Vehicle Detection and Classification System for Real-time Traffic Data Collection Using Uncalibrated Video Cameras. Transportation Research Record, No. 1993, pp. 138--147. TRB, Washington, D.C. (2007)
10. Zhang, W., Wu, J., Yin, H.: Moving vehicles detection based on adaptive motion histogram. Digital Signal Process., doi:10.1016/j.dsp.2009.10.006, pp. 1--13. Elsevier Inc. (2009)